# Enhancing Clinical Predictive Modeling through Model Complexity-Driven Class Proportion Tuning for Class Imbalanced Data: An Empirical Study on Opioid Overdose Prediction

Yinan Liu[1], Xinyu Dong[1], Weimin Lyu[1], MS, Richard N. Rosenthal[1], MD, Rachel Wong[1], MD, Tengfei Ma[2], Ph.D., Fusheng Wang[1], Ph.D.
[1]Stony Brook University, Stony Brook, NY
[2]IBM Research, Yorktown Heights, NY

**Abstract**

*Class imbalance problems widely exist in the medical field and heavily deteriorates performance of clinical predictive models. Most techniques to alleviate the problem rebalance class proportions and they predominantly assume the rebalanced proportions should be a function of the original data and oblivious to the model one uses. This work challenges this prevailing assumption and proposes that links the optimal class proportions to the model complexity, thereby tuning the class proportions per model. Our experiments on the opioid overdose prediction problem highlights the performance gain of tuning class proportions. Rigorous regression analysis also confirms the advantages of the theoretical framework proposed and the statistically significant correlation between the hyperparameters controlling the model complexity and the optimal class proportions.*

## Introduction

The class imbalance problem often occurs in medical machine learning because "positive" patients generally comprise a small fraction of the total population. For example, the positive rate is less than 1% for opioid overdose, about 2% for opioid use disorder, and between 3% and 20% for acute kidney injury in the Health Facts database.[22, 23, 24] In these settings, we may use metrics such as F1 score, precision, and recall to evaluate model performance, since classification errors can mislead, e.g., a classifier always predicting negative already has a low classification error.

Optimizing an F1 score is usually done by manipulating the input distribution without changing the internals of machine learning algorithms 10 . For instance, to "encourage" the classifier to pay more attention to positive instances, we can impose a heavier penalty on the misclassification of positive instances (hereafter "cost-sensitive learning") or replicate more positive instances in the training data (hereafter "resampling techniques"). Techniques such as SMOTE, ADASYN, and GAN family methods [3, 5, 19, 20] are based on similar principles. For both cost-sensitive learning and resampling techniques, we only need to tune one hyper-parameter, the positive weight scalar (PWS): in cost-sensitive learning we use the scalar to amplify the cost term for each positive instance, and in resampling techniques we use it to increase the probability that a positive instance is sampled. Studies on the determination of this hyper-parameter [3, 8, 9, 12, 14, 17, 21] assert that PWS should merely be a function of the structure of data. Let $\rho$ be the ratio between negative and positive instances. PWS as inverse class frequency $\rho$ or inverse square root of class frequency $\sqrt{\rho}$ are two common approaches. Referring to the opioid abuse detection problem, where only 1% of the instances are positive, we replicate the positive instances 10 (using the $\sqrt{\rho}$ rule) or 100 (using the $\rho$ rule) times in training.

In this paper, we formulate the following hypotheses: *H1. Tradeoffs across different error metrics*: PWS only provides tradeoffs between an F1 score and other error metrics and does not produce strictly better models. *H2. Tradeoffs between bias and variance in F1 score optimizations*: A larger PWS will usually improve an in-sample F1 score, reduce the effective sample, and degrade test performance.

*Validation/our methodology*

We start with a small synthetic study to highlight that tweaking the input distribution or cost function only results in a tradeoff, e.g., improving F1 scores at the cost of other metrics, and does not produce a strictly better model. We then analyze the role of PWS in predicting opioid overdose by training a massive number of models with different hyper-parameters in four widely used model families, perform regression analysis to understand the relationship between optimal PWS and model complexity, and confirm our hypotheses.

*Theoretical and practical implications*

Theoretically, our findings radically depart from the current ML techniques used to solve class imbalance problems. F1 score represents a balanced measurement for precision and recall and is preferred when the data is unbalanced [31]. The variance-bias tradeoff in optimizing F1 scores is clarified by linking optimal class proportions to the model complexity. Practically, our findings suggest that when positive weight scalar (PWS) is included, most off-the-shelf models will achieve considerable F1 score improvements with near-zero cost adjustments of the hyperparameter search. Prior to us, medical machine learning research rely more heavily on established ML libraries than designing new specialized models (primarily because of the cost) and hyperparameter search is usually restricted by available compute resources and routinely overlooks PWS, leading to suboptimal models.

Figure 1 shows that reweighing the training data does not produce better models.

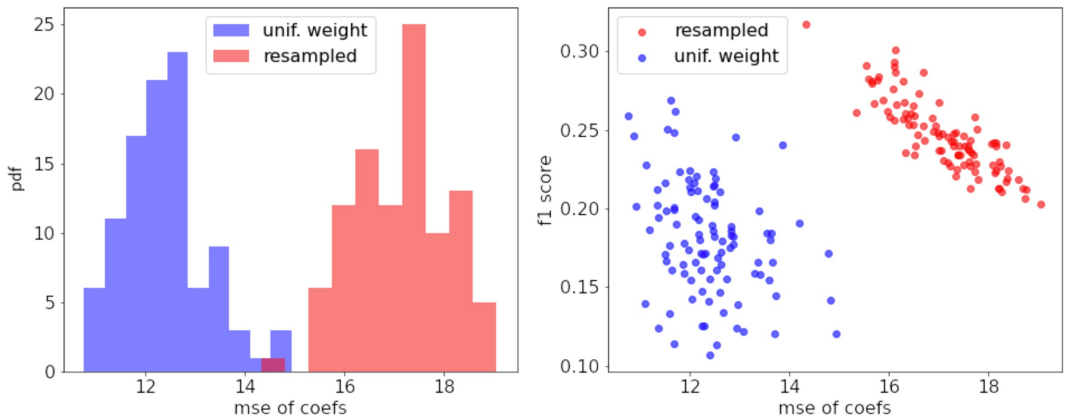

**Figure 1**. *Reweighting the training data only results in a tradeoff and does not produce better models.* PWS only results in a different tradeoff for a model (better F1 and worse MSE in this case) and does not produce strictly better predictions. We use a logistic linear model to generate synthetic training data under $D_0$ (blue) and $D_1$ (red). Only 5% of data sampled from $D_0$ are positive, whereas around 50% of data from $D_1$ are positive. We then use the datasets to fit logistic models. Left: Histogram of $l_2$-error of the estimated coefficients compared to ground-truth shows that models using data under $D_1$ (more balanced data) have worse performance. Right: Scatter plots between F1 scores and MSE of the estimated coefficients show that models using balanced data have worse MSE but improved F1 scores.

**Preliminary**

*Problem formulation*

We consider a binary classification problem using $d$-dimensional features. Our goal is to fit a function

$$y \sim f(x),$$

where $y \in \{0, 1\}$ and $x \in R^d$. We let $f(\cdot)$ be the commonly used machine learning black boxes such as gradient boosted decision trees [4, 11, 13, 15, 16], or neural networks[7], and let $T = \{(x_i, y_i)\}_{i \leq n}$ be the set of training data. The class imbalance problem refers to that the fraction of positive labels is substantially smaller than the fraction of negative labels; under this setting more robust metrics such as F1 score are often used to replace mean square errors (MSEs) or classification errors to assess model performance. We define F1 score as:

$$F1\ score = \frac{TP}{TP + \frac{1}{2}(FP + TN)},$$

where $TP$ is the number of true positives, $FP$ is the number of false positives, and $FN$ is the number of false negatives. We note that precision and recalls can also define F1 score [31].

*Training and addressing class imbalance problems.* Many tools and algorithms generally use existing ML architecture and tweak the cost function or the distribution of the training data to incentivize classifiers to err less on positive instances and improve F1 scores. Specifically, we let $l(y_i, \hat{y}_i)$ be a loss function used in an ML model, e.g., $l(y, \hat{y}) = (y - \hat{y})^2$ for MSE and $l(y, \hat{y}) = -(y \cdot log(\hat{y}) + (1 - y) \cdot log(1 - \hat{y}))$ for binary cross entropy loss. Without the presence of class imbalance problems, typically we would minimize

$$\sum_{(x_i, y_i) \in T} l(y_i, f(x_i)) = \sum_{(x_i, y_i) \in T} l(y_i, \hat{y}_i).$$

(1)

Cost sensitive learning and resampling are often used to address the class imbalance problem, both of which aim to increase the cost of making errors on positive instances and are equivalent under certain conditions CITE. The use of cost sensitive learning increases the cost for the positive instances, i.e., the new cost function becomes

$$\sum_{(x_i,y_i)\in T_P} \gamma \cdot l(y_i, f(x_i)) + \sum_{(x_i,y_i)\in T_N} l(y_i, f(x_i)),$$
(2)

where $T_P = \{(x_i, y_i) \in T | y_i = 1\}$ is the subset of positive instances in $T$, $T_N = T - T_P$ is the subset of negative instances, and $\gamma > 1$ is PWS. The use of resampling replaces the training set used in *Eq.1* to $D$ (i.e., we optimize $\sum_{(x_i,y_i)\in D} l(y_i, f(x_i))$, where each element in $D$ is sampled from $T$ so that $(x_i, y_i)$ is sampled with probability $p$ when $y_i = 0$ and $\gamma \cdot p$ when $y_i = 1$. Here, $\gamma$ is PWS and $p$ is a normalizing constant so that all probability masses sum to 1.

**Methodology and Experimental Design**

This section describes the setup of our experiments to confirm *H1* and *H2*. We first examine a synthetic dataset to gain intuitions. Then we explain the experiments for the opioid overdose problem based on real data.

*Intuition of error metrics' tradeoff*

As mentioned, we start with a small synthetic study and use a synthetic dataset produced by a logistic regression model to understand the relationship between PWS and model complexity. Recall that in a logistic model,

$$Pr[y_i = 1 | x_i] = \sigma(\omega^\intercal x_i + \omega_0).$$
(3)

In a standard convergence analysis, we assume that i.i.d. samples $(x_i, y_i)$ from distribution $D$ are observed and we find an estimator $(\hat{\omega}, \hat{\omega}_0)$ that optimizes the likelihood function. The maximum-likelihood estimator (MLE) $(\hat{\omega}, \hat{\omega}_0)$ is unbiased and its variance is usually related to the covariance matrix of $x_i$ (i.e., asymptotic normality of MLE [18]).

We consider two distributions $D_0$ and $D_1$ in which $D_0$ is the original distribution that results in imbalanced $y$'s, whereas $D_1$ tweaks the distribution for $x$ so that the fraction of positive $y$'s become larger. We consider training two models using the same amount of data, where the MLEs constructed under both distributions are asymptotically normal, i.e., both have the same mean that coincides with the ground-truth, and different variances. Because two estimators are trained using the same number of samples, it is unlikely that the quality of one estimator predominates, i.e., the confidence interval (set) or the equidistant contours are only rotated but not shrunk when the data distribution changes from $D_0$ to $D_1$.

To illustrate, we let $d = 100$ (number of features) and $n = 120$ (number of observations) and use the model specified in *Eq.3* to generate the training data. We set $\omega_0$ to be a negative number with a suitable magnitude so that the fraction of positive instances is small when $E[x] = 0$. For $D_0$, we assume that $x$ follows standard normal distribution. For $D_1$, we follow $D_0$ to sample a sufficiently large training data and then delete a large fraction of negative instances to obtain an approximate balance between the positive and negative instances. We note that $D_1$ emulates the standard over-sampling trick.

We generate 100 groups of data for each of $D_0$ and $D_1$ and examine the distributions of estimators. Referring to Fig. 1, we see that the estimators from $D_1$ are worse than those from $D_0$ if we measure the MSE of the estimators (Fig. 1 left), but the F1 scores are higher under $D_1$. In Fig. 1 (right), we see that the scatter points of models under $D_1$ moving from lower right to upper left indicate the tradeoff, i.e., F1 improves at the cost of estimator accuracy.

Results of this synthetic study have helped us to construct *H2*: a large PWS rotates the confidence set to a direction that is better for F1 score but it also reduces the effective sample size. Consider for example the case PWS → ∞. Here we care effectively only about the positive instances and the training set size is substantially smaller. Thus, we shall expect to see a tradeoff: as the complexity of the models grows we first need to scale up PWS, but at a certain point PWS needs to stop growing and shrink to prevent the volume of confidence set to grow excessively large due to lack of effective samples. This also implies PWS should be optimized per model instead of being set as a constant.

*Dataset*

We adopt data from Cerner's Health Facts,[22] a major EHR database in the United States. It has information of nearly 69 million patients from over 600 participating clinical facilities. This database is structured based on patients' encounters, which contain diagnoses, procedures, patients' demographics, medication dosage and administration information, vital signs, laboratory test results, surgical case information, health systems attributes and other clinical observations. For this opioid overdose study, we extracted all patients with opioid prescriptions from Health Facts. After extraction, there are 5,231,614 patients left, 44,774 of them are positive cases. The positive rate is less than 1%. See also Fig. 2 for our data collection workflow.

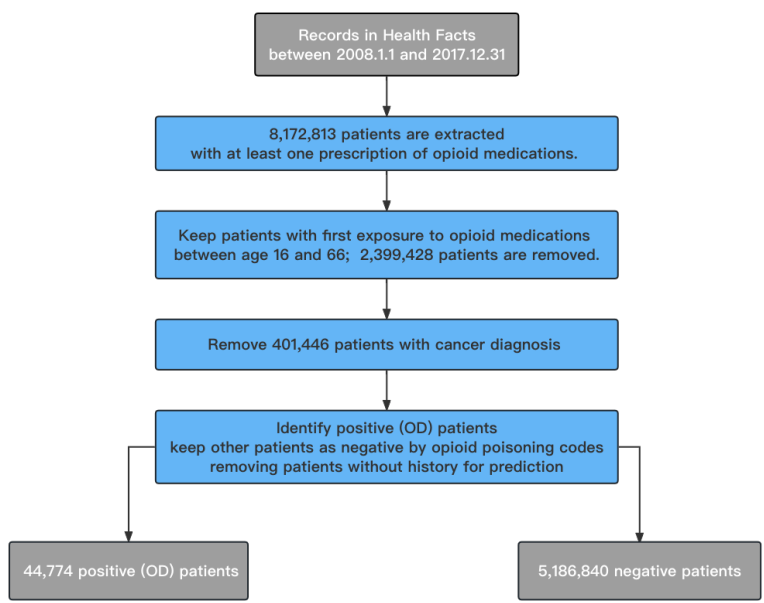

**Figure 2**. A flowchart of collecting and generating experimental data

To build the feature input from the dataset, we followed predefined process[22] to extract 1185 features, including 414 diagnosis codes features, 394 laboratory test features, 3 demographic features, 227 clinical events features, and 147 medications features. With the constructed feature matrix, we apply the BERT model to be trained on it by the two classic pretraining ways including masked language model and next sentence prediction[33]. After training, the output of each patient input will be used as a patient presentation of 50 dimensions. We used the representations to conduct our research.

*Models*

We select four machine learning models widely used in applied machine learning research that routinely outcompete more sophisticated and/or customized models in public benchmarks[2, 11, 15]. Two models are based on boosted classification trees (lightgbm and catboost), and the other two are based on deep learning. We train 8000 lightgbm models, 2000 catboost models, and 16000 neural net models for a total of 26,000 models. For each model, we search between 1 and 150 to find the optimal PWS. Table 1 lists the models, hyper-parameters, and values.

**Table 1**. Four machine learning models and the hyper-parameters and values examined.

| Model name | Hyper-parameter | Values of hyper-parameters |
|---|---|---|
| Lightgbm | Number of estimators (trees) | 100, 500, 1000, 2000, 5000, 8000, 10000 |
|  | Max depth of a tree | 4, 5, 7 |
|  | Max number of leaves in a tree | 100, 300, 500, 1000 |
|  | Learning rates | 0.1, 0.05, 0.01 |
| Catboost | Number of estimators (trees) | 100, 500, 1000, 2000, 5000, 8000, 10000 |
|  | Max depth of a tree | 4, 5, 7 |
|  | Learning rates | 0.1, 0.05, 0.01 |

| MLP (both weighted cost and resampling) | Width of the hidden layer(s) | 50, 100 |
|---|---|---|
| | Number of layers | 1, 2 |
| | Scale down initialization | 1, 0.1 |
| | Batch normalization | No, only after input, after every linear-relu layer |
| | Learning rates | $10^{-1}, 10^{-2}, 10^{-3}, 10^{-4}$ |
| | Momentum | 0.1, 0.9 |

*Lightgbm* [11]

The Lightgbm model offers a set of highly effective heuristic learning rules while maintaining comparable computational efficiency. It implements a variant of gradient boosted regression trees and progressively builds a collection of tree-based base learners. Each base learner is a decision tree with its depth and number of leaves controlled by hyper-parameters. After training a new base learner, the model adjusts the response by residualizing it against the estimation from the new base learner.

*Catboost* [15]

The Catboost model implements another variant of gradient boosted decision trees. It integrates a larger number of new heuristics and statistical estimation techniques in training. The new techniques properly address a so-called "target leak" problem found in boosting-based models.

*Neural nets weighted cost function*

We consider multi-layer perceptrons with one or two hidden layers. We wire PWS through the cost function and optimize the cost function from *Eq.2*.

*Neural nets resampling*

We use the same set of model architecture as above. We wire PWS through resampling, i.e., a positive instance is $\gamma$ times more likely to be sampled than a negative instance in a stochastic gradient batch, where $\gamma$ is PWS.

**Model complexity**

We next describe the hyperparameters that can potentially relate to the complexity of the models which will be used to perform satisfied analysis between model complexity and optimal PWS. *(i)* Lightgbm and Catboost: Both the number of leaves, the depth of the tree, and learning rates could be related to these models' complexity. Number of leaves is also related to lightgbm's model complexity (catboost cannot tune this hyper-parameter). *(ii)* Neural nets: The width of each layer and the number of layers is most related to model complexity.

In general, it is not always straightforward to determine whether a hyperparameter is related to model complexity. *(i)* we determine that the learning rate, batch size, momentum, or scale of randomly initialized learnable parameters are not part of neural nets' model complexity although some evidence suggests they could potentially regularize fitting[32]. *(ii)* while neural nets with weighted cost functions and those with resampling share the same architecture, they seem to have different capacity in fitting training data so they may not have the same model complexity. *(iii)* Catboost can still fit training data well even with a small number of trees/estimators, suggesting that it may be harder to characterize catboost's model complexity. We shall see in the next section that quantifying a model's complexity is indeed a delicate exercise.

We consider multi-layer perceptrons with one or two hidden layers. We train these neural nets in two ways. *(i)* Using weight-adjusted cost function. We wire PWS through the cost function and optimize the cost function from *Eq.2*. *(ii)* Using resampling. We wire PWS through resampling, i.e., a positive instance is $\gamma$ times more likely to be sampled than a negative instance in a stochastic gradient batch, where $\gamma$ is PWS.

Model complexity of neural nets. The size of neural networks (e.g., the width of each layer and the total number of layers) are most related to the model complexity. The landscape in practice is more delicate. For example, while two families of neural nets (NN with weight adjusted cost and NN with re-sampling) theoretically minimize the same utility function, it seems NN with resampling is better at fitting training data so it may have higher model complexity.

## Experimental results

This section describes the results of our experiments and our regression analyses. We first examine the performance improvements obtained by tuning PWS for all four families of models with different hyper-parameters. Our baseline models assume that PWS is a constant set to 10 or 100. Table 2 lists the results.

**Table 2.** Performance of models with different PWS. We considered lightgbm (lgbm), catboost, multi-layer perceptron with weighted cost (mlp.wcost), and multi-layer perceptron with resampling (mlp.wsample). We also consider three ways of determining PWS, including grid search (rows with suffix '.opt'), using the linear rule to set it as constant 100 (rows with suffix 'w.100'), and using the square root rule to set it as constant 10 (rows with suffix 'w.10'). We display f1 score and error for training and test, and roc, precision, and recall for test. The PWS column is the optimal PWS for the ones using grid search, and the constants used for the one using linear or square root rules.

| model | f1_test | PWS | roc_test | precision_test | recall_test | error_test | f1_train | error_train |
|---|---|---|---|---|---|---|---|---|
| **lgbm.opt** | **53.1%** | **25** | 95.1% | 57.8% | 49.2% | 1.5% | 49.1% | 1.5% |
| Baseline: lgbm.fixed.w | 51.5% | 10 | 95.0% | 57.3% | 47.1% | 1.4% | 53.2% | 1.3% |
| Baseline: lgbm.fixed.w | 51.3% | 100 | 95.1% | 59.3% | 45.4% | 6.5% | 48.6% | 6.5% |
| **catboost.opt** | **52.1%** | **35** | 94.4% | 58.6% | 47.1% | 2.8% | 57.2% | 2.5% |
| Baseline: cb.fixed.w | 51.0% | 10 | 95.1% | 57.5% | 45.9% | 1.4% | 52.2% | 1.2% |
| Baseline: cb.fixed.w | 50.4% | 100 | 93.8% | 56.9% | 45.3% | 4.4% | 53.7% | 4.1% |
| **mlp.wcost.opt** | **50.8%** | **70** | 86.9% | 55.4% | 47.2% | 4.9% | 48.7% | 4.8% |
| Baseline: mlp.wcost.10 | 50.3% | 10 | 82.4% | 51.2% | 49.8% | 2.2% | 48.8% | 2.2% |
| Baseline: mlp.wcost.100 | 50.4% | 100 | 90.3% | 54.0% | 47.4% | 7.8% | 48.7% | 7.8% |
| **mlp.wsample.opt** | **50.8%** | **57** | 86.4% | 52.1% | 49.6% | 5.6% | 49.1% | 5.5% |
| Baseline: mlp.wsample.10 | 50.5% | 10 | 79.3% | 56.4% | 45.8% | 1.4% | 50.2% | 1.4% |
| Baseline: mlp.wsample.100 | 50.3% | 100 | 90.8% | 53.3% | 47.9% | 8.7% | 48.7% | 8.6% |

We have three major finding results. First, tuning only PWS improves the F1 scores in all scenarios. Since the models tend to plateau at around 50% [6], a 53.1% F1score for lgbm.opt denotes a significant improvement. Second, model-dependent optimal PWS is consistent with H2. Third, the tradeoffs (F1 scores improve but classification errors increase) between the two-performance metrics also confirm *H1*.

We also produce scatter plots of the F1 scores for the four families of models. In Figure 2, each point corresponds to one model with a fixed hyper-parameter set, the $x$-axis represents the models' F1 scores for the training data, and the y-axis represents the F1 scores for the test data.

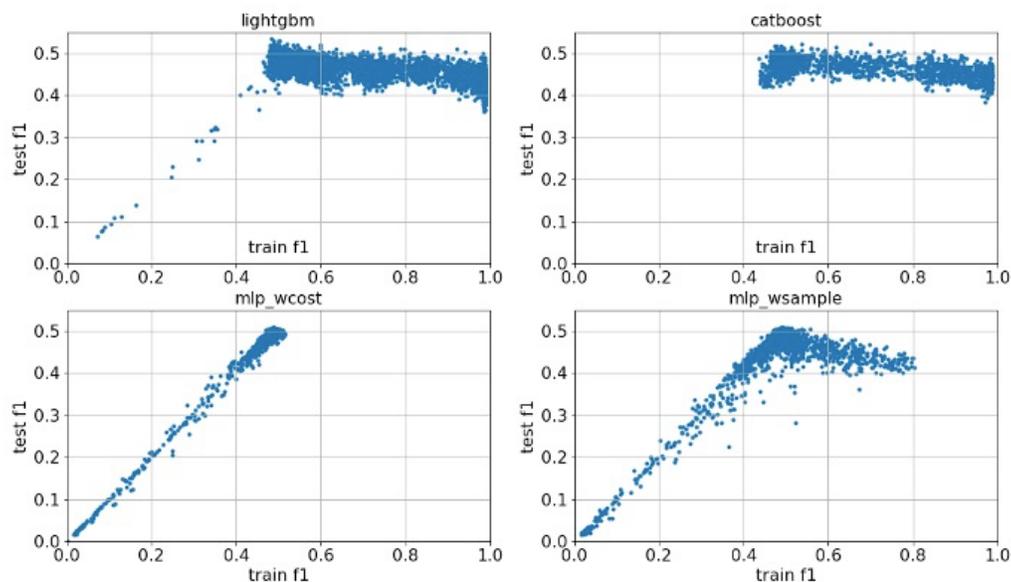

**Figure 3.** Scatter plots between training and test F1 scores for lightgbm, catboost, multi-layer perceptron with weighted cost (mlp_wcost), and multi-layer perceptron with resampling (mlp_wsample).

We find three major results. First, all four families have variance-bias tradeoff phenomena. When the models' training F1 scores improve, the test F1 scores improve until the training and test F1 scores are around 50%. Afterward, while the training F1 scores can continue to improve, the test F1 scores either stay flat (catboost models) or reduce (lightgbm and MLP with resampling). Second, while lightgbm and catboost are variants of gradient boosted regression trees, they have substantially different training behaviors, i.e., all catboost models fit the training data well (have $\geq 0.4$ F1 scores), whereas a considerable portion lightgbm models do not properly fit the training data (i.e., $\leq 0.4$ F1 scores). Third, all MLP with weighted cost functions fail to fit the training data (have $\leq 0.6$ scores), whereas MLP with resampling fit the training data better. Note that these two families of neural nets use the same set of architectures. All these findings suggest that characterizing model complexity via hyper-parameters is remarkably challenging.

**Linear regression analyses**

This section analyzes the relationship between PWS and model complexities based on the 26,000 models we trained. We use linear regression because it has more robust procedures to estimate model coefficients and the $t$-statistics, although model complexity variables and PWS may not be exactly linearly related. We start with lightgbm, because its statistical behaviors are well known and it requires few heuristics, followed by the catboost and neural net models.

*Lightgbm regression model*

As mentioned earlier, the hyper-parameters related to lightgbm's complexity include the number of estimators, learning rates, maximum depth for each tree, and maximum number of leaves in each tree. For each set of hyper-parameters, we find $k$ best PWS values. Then we construct $k$ observations, each using model hyper-parameter as covariates and optimal PWS as responses. We then fit the following model:

$$y \sim \beta_0 + \beta_1 x_1 + \beta_2 x_2 + \beta_3 x_3 + \beta_4 x_4 + \epsilon$$

where $x_1$ is the number of estimators, $x_2$ is learning rates, $x_3$ is max tree depth, $x_4$ is the number of leaves, and the response $y$ is the corresponding optimal PWS. We fit for both training and test data (i.e., choose models with the best training F1 scores for training data and choose those with best test scores for test data). The number of PWS's for each hyper-parameter group $k$ controls a tradeoff between estimation accuracy and estimation quality. When $k = 1$, we focus only on the optimal PWS and obtain the most ideal response, but the total number of observations is smaller. When $k$ becomes larger, we pick the top $k$ models for each set of hyper-parameters and note that even though some non-best models are selected, the total number of observations is larger. Multiple $k$'s is examined to improve the robustness of our analysis.

From *H2*, we expect PWS will eventually need to shrink for test performance as model complexity increases. In addition, the optimal PWS for training data should increase as more complex models are used. Specifically, when we use simple models, the set of decision boundaries is quite rigid (i.e., decision boundaries for linear models are linear). In this case, large PWS could alter the decision boundaries to undesirable positions, thus harming the F1 scores of both the training and test data. When more complex models are used, decision boundaries have higher degrees of freedom so increasing PWS improve training F1 scores.

Table 3 lists the estimated coefficient, the $t$-statistics associated with the coefficient and the correlations between each covariate and the response. The $t$-statistics (whether the relationships are statistically significant) and their directions are the most important. We consider a variable statistically significant when the absolute value of its $t$-statistics is at least 3 (corresponding to $p$-value 0.003).

**Table 3.** Regression analysis for lightgbm

| k | mode | counts | n_estimaters ($x_1$) | | | learning_rate ($x_2$) | | | max_depth ($x_3$) | | | num_leaves($x_4$) | | |
|---|---|---|---|---|---|---|---|---|---|---|---|---|---|---|
| | | | beta-$\beta_1$ | t-stat | corr | beta-$\beta_2$ | t-stat | corr | beta-$\beta_3$ | t-stat | corr | beta-$\beta_4$ | t-stat | corr |
| 1 | test | 252 | -0.0027 | -3.4665 | -0.2138 | -62.6178 | -0.8179 | -0.0505 | -3.9539 | -1.7608 | -0.1086 | 0.0019 | 0.2269 | 0.0140 |
| | train | 252 | 0.0062 | 10.5165 | 0.5177 | 384.4424 | 6.6113 | 0.3255 | 5.7409 | 3.3444 | 0.1646 | -0.0026 | -0.4086 | -0.0201 |
| 3 | test | 756 | -0.0024 | -5.4263 | -0.1933 | -18.9180 | -0.4278 | -0.0152 | -3.5534 | -2.7223 | -0.0970 | -0.0028 | -0.5710 | -0.0203 |
| | train | 756 | 0.0044 | 13.5818 | 0.4128 | 342.6457 | 10.6497 | 0.3237 | 5.4452 | 5.7331 | 0.1743 | -0.0029 | -0.8238 | -0.0250 |
| 5 | test | 1260 | -0.0021 | -5.9845 | -0.1660 | -15.6023 | -0.4524 | -0.0125 | -3.0111 | -2.9575 | -0.0820 | -0.0003 | -0.0739 | -0.0021 |
| | train | 1260 | 0.0040 | 15.5481 | 0.3781 | 307.4436 | 12.0934 | 0.2941 | 5.1650 | 6.8824 | 0.1674 | 0.0017 | 0.6099 | 0.0148 |

<u>Discussion.</u> We find the results are consistent with our expectations from H2. First, $\beta_1$ to $\beta_4$ are positive in the regression models for the training data, which suggests that PWS increases to optimize F1 scores as the models grow more complex. We find negative $\beta_1$, $\beta_2$, and $\beta_4$ in the regression models for the test data, which suggests that PWS decreases to optimize F1 scores as the models grow more complex and find that $\beta_3$ (the coefficient associated with max depth) is similar, although less statistically significant.

*Catboost regression model*

We mimic lightgbm, but since we cannot directly control the number of leaves in the hyper-parameters in catboost, our regression becomes:

$$y \sim \beta_0 + \beta_1 x_1 + \beta_2 x_2 + \beta_3 x_3 + \epsilon,$$

where $x_1$ is the number of estimators, $x_2$ is learning rates, $x_3$ is max tree depth, and $y$ is the corresponding PWS. We continue to examine $k = 1, 3, 5$ and run the model for training and test sets. Again, we note that measuring the complexity of catboost models can be subtle because even when a catboost model consists of a small number of trees (estimators), it can still fit the training data well. To address this issue, we remove catboost models with small numbers of trees from our regression analysis. Table 4 lists the results for catboost models with 2,000 or more trees.

**Table 4.** Regression analysis for catboost with 2,000 or more trees

| k | mode | n | n_estimaters ($x_1$) | | | learning_rate ($x_2$) | | | max_depth ($x_3$) | | |
|---|---|---|---|---|---|---|---|---|---|---|---|
| | | | beta-$\beta_1$ | t-stat | corr | beta-$\beta_2$ | t-stat | corr | beta-$\beta_3$ | t-stat | corr |
| 1 | test | 36 | 0.0033 | 1.3943 | 0.2051 | 520.8121 | 2.6380 | 0.3881 | -13.4259 | -2.3037 | -0.3389 |
| | train | 36 | 0.0006 | 2.1468 | 0.2643 | 88.7907 | 3.8464 | 0.4736 | 2.5995 | 3.8147 | 0.4697 |
| 3 | test | 108 | 0.0022 | 1.4830 | 0.1414 | 230.4582 | 1.8909 | 0.1803 | -1.6923 | -0.4704 | -0.0448 |
| | train | 108 | 0.0005 | 1.8348 | 0.1596 | 73.6998 | 3.2451 | 0.2822 | 2.5384 | 3.7862 | 0.3293 |
| 5 | test | 180 | 0.0018 | 1.5067 | 0.1122 | 120.7273 | 1.2368 | 0.0921 | -1.8289 | -0.6347 | -0.0473 |
| | train | 80 | 0.0004 | 1.4496 | 0.1027 | 58.3983 | 2.8842 | 0.2044 | 2.1347 | 3.5715 | 0.2531 |

*Discussion.* The results are similar to lightgbm. We find mostly positive $\beta_i$ for the training data. The coefficients associated with learning rates and max depth are significant in training, whereas all coefficients are insignificant (from 0, i.e., we cannot reject the null hypothesis that all coefficients are 0) for the test data. Unlike lightgbm, the coefficients associated with the number of estimators and max depth are positive (albeit not significant) in test set, suggesting that catboost is better at preventing overfitting. In general, it also reflects that measuring model complexity for more recent models that use advanced statistical techniques is harder.

*MLP regression models*

Compared to boosting-based models, MLP regression analysis requires examining a larger number of models for each set of complexity-related hyper-parameters. In addition to tuning the width and number of hidden layers, we also need to tune many other hyperparameters (i.e., random initialization scales, whether to batch normalize, learning rates, and momentum) to optimize training performance. To address this issue, we remove irrelevant hyper-parameters from our regression analyses.

Our final regression model consists of only two covariates:

$$y \sim \beta_0 + \beta_1 x_1 + \beta_2 x_2 + \epsilon,$$

where $x_1$ is the width of each hidden layer, and $x_2$ is the number of hidden layers. We examine 1,000 models check, and increase $k$ (i.e., $k = 100, 200, 300$) to obtain more robust responses. Table 5 lists the results.

**Table 5.** Regression analysis for MLP

| k | mode | n | hidden_size | | | n_layers | | |
|---|---|---|---|---|---|---|---|---|
| | | | beta-$\beta_1$ | t-stat | corr | beta-$\beta_2$ | t-stat | corr |
| 100 | test | 800 | -0.0016 | -0.3696 | -0.0131 | -3.1375 | -0.9698 | -0.0343 |
| | train | 800 | 0.0324 | 8.8693 | 0.2720 | 38.1875 | 13.6954 | 0.4200 |
| 200 | test | 1600 | 0.0010 | 0.3371 | 0.0084 | -1.1750 | -0.5111 | -0.0128 |
| | train | 1600 | 0.0174 | 5.9822 | 0.1457 | 16.1938 | 7.3153 | 0.1781 |
| 300 | test | 2400 | -0.0006 | -0.2439 | -0.0050 | 0.1083 | 0.0577 | 0.0012 |
| | train | 2400 | 0.0079 | 3.3094 | 0.0673 | 5.3292 | 2.9400 | 0.0598 |

*Discussion.* We find that MLP resembles the lightgbm and catboost results. We find that $\beta_1$ and $\beta_2$ are positive and significant in the training set, whereas all coefficients are insignificant in the test set, which again suggests the presence of two competing forces (improving F1 score vs reducing variance error) controls the optimal values of PWS.

As we have noted, since the width and number of hidden layers may not properly characterize the complexity of deep neural nets, it may be the reason that we do not see $\beta_1$ and $\beta_2$ being negative and significant for test sets. Therefore, we need to find other covariates to characterize model complexity. A natural statistic is a model's *ensample* performance, i.e., when a model fits the training data better, it exhibits stronger fitting power and is more likely to be complex. Thus, we add the ensample F1 performance in our regression model, and estimate:

$$y \sim \beta_0 + \beta_1 x_1 + \beta_2 x_2 + \beta_3 x_3 + \epsilon,$$

where $x_1$ is the width of each hidden layer, $x_2$ is the number of hidden layers, and $x_3$ is the in-sample F1 score. Table 6 shows the results.

**Table 6**. Regression analysis for MLP with ensample covariate

| k | mode | n | hidden_size beta-$\beta_1$ | t-stat | corr | n_layers beta-$\beta_2$ | t-stat | corr | f1_train beta-$\beta_3$ | t-stat | corr |
|---|---|---|---|---|---|---|---|---|---|---|---|
| 100 | test | 800 | 0.0087 | 2.2328 | -0.0131 | 3.7541 | 1.2726 | -0.0343 | -2,469.1991 | -13.8136 | -0.4335 |
| 200 | test | 1600 | 0.0081 | 2.9607 | 0.0084 | 4.2574 | 2.0507 | -0.0128 | -2,859.9630 | -19.9198 | -0.4392 |
| 300 | test | 2400 | 0.0066 | 2.9997 | -0.0050 | 5.8308 | 3.4562 | 0.0012 | -3,022.3433 | -25.1486 | -0.4495 |

*Discussion.* We find that $\beta_3$ has a very strong negative effect on the optimal PWS. In other words when a model fits the in-sample data better, it needs a smaller PWS to better perform in test data. This finding confirms H2.

**Conclusion**

This work revisits the widely used rules to determine class proportions in the rebalancing process for tackling class imbalance problems. Our work proposed to link the optimal class proportions/positive weight scalar (PWS) to the complexity of clinical predictive models used to solve class imbalance problems. Experiments on the opioid overdose problem revealed visible improvements in the proposed framework's predictive power. In addition, regression analysis found a statistically significant correlation between the hyperparameters controlling model complexity and PWS. The theoretical framework may be applied to solve a variety of class imbalance problems and improve predictive performance at near-zero cost.

**References**

1. Kayley Abell-Hart, Sina Rashidian, Dejun Teng, Richard N Rosenthal, and Fusheng Wang. Where opioid overdose patients live far from treatment: Geospatial analysis of underserved populations in new york state. JMIR Public Health and Surveillance, 8(4):e32133, 2022.
2. Casper Solheim Bojer and Jens Peder Meldgaard. Kaggle forecasting competitions: An overlooked learning opportunity. International Journal of Forecasting, 37(2):587–603, 2021.
3. Nitesh V Chawla, Kevin W Bowyer, Lawrence O Hall, and W Philip Kegelmeyer. Smote: synthetic minority over-sampling technique. Journal of artificial intelligence research, 16:321–357, 2002.
4. Tianqi Chen, Tong He, Michael Benesty, Vadim Khotilovich, Yuan Tang, Hyunsu Cho, Kailong Chen, Rory Mitchell, Ignacio Cano, Tianyi Zhou, et al. Xgboost: extreme gradient boosting. R package version 0.4-2, 1(4):1–4, 2015.
5. Yin Cui, Menglin Jia, Tsung-Yi Lin, Yang Song, and Serge Belongie. Class-balanced loss based on effective number of samples. In Proceedings of the IEEE/CVF conference on computer vision and pattern recognition, pages 9268–9277, 2019.
6. Xinyu Dong, Rachel Wong, Weimin Lyu, Kayley Abell-Hart, Jianyuan Deng, Yinan Liu, Janos G Hajagos, Richard N Rosenthal, Chao Chen, and Fusheng Wang. An integrated lstm-heterorgnn model for interpretable opioid overdose risk prediction. Artificial intelligence in medicine, 135:102439, 2023.
7. Ian Goodfellow, Yoshua Bengio, and Aaron Courville. Deep learning. MIT press, 2016.
8. Haibo He, Yang Bai, Edwardo A Garcia, and Shutao Li. Adasyn: Adaptive synthetic sampling approach for imbalanced learning. In 2008 IEEE international joint conference on neural networks (IEEE world congress on computational intelligence), pages 1322–1328. IEEE, 2008.
9. Chen Huang, Yining Li, Chen Change Loy, and Xiaoou Tang. Learning deep representation for imbalanced classification. In Proceedings of the IEEE conference on computer vision and pattern recognition, pages 5375–5384, 2016.
10. Nathalie Japkowicz and Shaju Stephen. The class imbalance problem: A systematic study. Intelligent data analysis, 6(5):429–449, 2002.
11. Guolin Ke, Qi Meng, Thomas Finley, Taifeng Wang, Wei Chen, Weidong Ma, Qiwei Ye, and Tie-Yan Liu. Lightgbm: A highly efficient gradient boosting decision tree. Advances in neural information processing systems, 30, 2017.
12. Dhruv Mahajan, Ross Girshick, Vignesh Ramanathan, Kaiming He, Manohar Paluri, Yixuan Li, Ashwin Bharambe, and Laurens Van Der Maaten. Exploring the limits of weakly supervised pretraining. In Proceedings of the European conference on computer vision (ECCV), pages 181–196, 2018.

13. Llew Mason, Jonathan Baxter, Peter Bartlett, and Marcus Frean. Boosting algorithms as gradient descent. Advances in neural information processing systems, 12, 1999.
14. Tomas Mikolov, Ilya Sutskever, Kai Chen, Greg S Corrado, and Jeff Dean. Distributed representations of words and phrases and their compositionality. Advances in neural information processing systems, 26, 2013.11
15. Liudmila Prokhorenkova, Gleb Gusev, Aleksandr Vorobev, Anna Veronika Dorogush, and Andrey Gulin. Catboost: unbiased boosting with categorical features. Advances in neural information processing systems, 31, 2018.
16. Robert E Schapire and Yoav Freund. Boosting: Foundations and algorithms. Kybernetes, 42(1):164–166, 2013.
17. Yu-Xiong Wang, Deva Ramanan, and Martial Hebert. Learning to model the tail. Advances in neural information processing systems, 30, 2017.
18. Larry Wasserman. All of statistics: a concise course in statistical inference, volume 26. Springer, 2004.
19. Hao Yang and Yun Zhou. Ida-gan: A novel imbalanced data augmentation gan. In 2020 25th International Conference on Pattern Recognition (ICPR), pages 8299–8305. IEEE, 2021.
20. Tingting Zhou, Wei Liu, Congyu Zhou, and Leiting Chen. Gan-based semi-supervised for imbalanced data classification. In 2018 4th International Conference on Information Management (ICIM), pages 17–21. IEEE, 2018.
21. Yang Zou, Zhiding Yu, BVK Kumar, and Jinsong Wang. Unsupervised domain adaptation for semantic segmentation via class-balanced self-training. In Proceedings of the European conference on computer vision (ECCV), pages 289–305, 2018.12
22. Dong X, Deng J, Hou W, Rashidian S, Rosenthal RN, Saltz M, Saltz JH, Wang F. Predicting opioid overdose risk of patients with opioid prescriptions using electronic health records based on temporal deep learning. Journal of biomedical informatics. 2021 Apr 1;116:103725.
23. Dong X, Deng J, Rashidian S, Abell-Hart K, Hou W, Rosenthal RN, Saltz M, Saltz JH, Wang F. Identifying risk of opioid use disorder for patients taking opioid medications with deep learning. Journal of the American Medical Informatics Association. 2021 Jul 30;28(8):1683-93.
24. Xu Z, Feng Y, Li Y, Srivastava A, Adekkanattu P, Ancker JS, Jiang G, Kiefer RC, Lee K, Pacheco JA, Rasmussen LV, Pathak J, Luo Y, Wang F. Predictive Modeling of the Risk of Acute Kidney Injury in Critical Care: A Systematic Investigation of The Class Imbalance Problem. AMIA Jt Summits Transl Sci Proc. 2019 May 6;2019:809-818. PMID: 31259038; PMCID: PMC6568062.
25. X. -Y. Liu, J. Wu and Z. -H. Zhou, "Exploratory Undersampling for Class-Imbalance Learning," in IEEE Transactions on Systems, Man, and Cybernetics, Part B (Cybernetics), vol. 39, no. 2, pp. 539-550, April 2009, doi: 10.1109/TSMCB.2008.2007853.
26. I. Fakhruzi, "An artificial neural network with bagging to address imbalance datasets on clinical prediction," 2018 International Conference on Information and Communications Technology (ICOIACT), Yogyakarta, Indonesia, 2018, pp. 895-898, doi: 10.1109/ICOIACT.2018.8350824
27. Sun S, Wang F, Rashidian S, Kurc T, Abell-Hart K, Hajagos J, Zhu W, Saltz M, Saltz J. Generating Longitudinal Synthetic EHR Data with Recurrent Autoencoders and Generative Adversarial Networks. InHeterogeneous Data Management, Polystores, and Analytics for Healthcare: VLDB Workshops, Poly 2021 and DMAH 2021, Virtual Event, August 20, 2021, Revised Selected Papers 7 2021 (pp. 153-165). Springer International Publishing.
28. van den Goorbergh R, van Smeden M, Timmerman D, Van Calster B. The harm of class imbalance corrections for risk prediction models: illustration and simulation using logistic regression. Journal of the American Medical Informatics Association. 2022 Sep;29(9):1525-34.
29. Vandewiele G, Dehaene I, Kovács G, Sterckx L, Janssens O, Ongenae F, De Backere F, De Turck F, Roelens K, Decruyenaere J, Van Hoecke S. Overly optimistic prediction results on imbalanced data: a case study of flaws and benefits when applying over-sampling. Artificial Intelligence in Medicine. 2021 Jan 1;111:101987.
30. Florence C, Luo F, Rice K. The economic burden of opioid use disorder and fatal opioid overdose in the United States, 2017. Drug Alcohol Depend. 2021 Jan 1;218:108350. doi: 10.1016/j.drugalcdep.2020.108350. Epub 2020 Oct 27. PMID: 33121867; PMCID: PMC8091480.
31. Sasaki Y. The truth of the f-measure. 2007. URL: https://www.cs.odu. edu/mukka/cs795sum09dm/Lecturenotes /Day3/F-measure-YS-26Oct07. pdf [accessed 2021-05-26].
32. Wu L, Li J, Wang Y, Meng Q, Qin T, Chen W, Zhang M, Liu TY. R-drop: Regularized dropout for neural networks. Advances in Neural Information Processing Systems. 2021 Dec 6;34:10890-905
33. Devlin J, Chang MW, Lee K, Toutanova K. Bert: Pre-training of deep bidirectional transformers for language understanding. arXiv preprint arXiv:1810.04805. 2018 Oct